%% file: main.tex
\documentclass[10pt,twocolumn,letterpaper]{article}

\usepackage[pagenumbers]{cvpr} 

\usepackage{graphicx}
\usepackage{amsmath}
\usepackage{amssymb}
\usepackage{booktabs}
\usepackage{nicefrac}
\usepackage{algorithm, algpseudocode}

\DeclareMathOperator*{\argmax}{arg\,max}

\algrenewcommand\algorithmicrequire{\textbf{Input:}}
\algrenewcommand\algorithmicensure{\textbf{Output:}}

\usepackage{listings}
\usepackage{xcolor,colortbl}
\definecolor{darkgreen}{rgb}{0.0, 0.5, 0.0}

\usepackage[pagebackref,breaklinks,colorlinks]{hyperref}

\usepackage[capitalize]{cleveref}
\crefname{section}{Sec.}{Secs.}
\Crefname{section}{Section}{Sections}
\Crefname{table}{Table}{Tables}
\crefname{table}{Tab.}{Tabs.}

\def\cvprPaperID{3438} 
\def\confName{CVPR}
\def\confYear{2023}

\begin{document}

\input{metadata}

\maketitle

\input{sec/0_teaser.tex}

\input{sec/1_abstract.tex}

\input{sec/2_intro.tex}
\input{sec/3_related_framefig.tex}
\input{sec/4_method.tex}
\input{sec/5_dataset.tex}
\input{sec/6_experiments.tex}
\input{sec/7_conclude.tex}

{\small
\bibliographystyle{ieee_fullname}
\bibliography{macros, references}
}

\newpage
\input{sec/8_appendix.tex}
\end{document}

%% file: metadata.tex
\title{Affective Faces for Goal-Driven Dyadic Communication}

\newcommand*\samethanks[1][\value{footnote}]{\footnotemark[#1]}

\author{Scott Geng\thanks{Equal contribution}\hspace{.2cm}
Revant Teotia\samethanks\hspace{.2cm}
Purva Tendulkar\hspace{.2cm}
Sachit Menon\hspace{.2cm}
Carl Vondrick\\[0.09cm]
Columbia University\\[0.01cm]
{\tt\footnotesize \{scott.geng, rt2819, pt2578, sachit.menon,  cv2428\}@columbia.edu}
}

%% file: sec/0_teaser.tex
\enlargethispage{-6.6cm} 
\noindent\begin{picture}(0,0)
\put(0,-403){\begin{minipage}{\textwidth}
\centering
\includegraphics[width=\linewidth]
{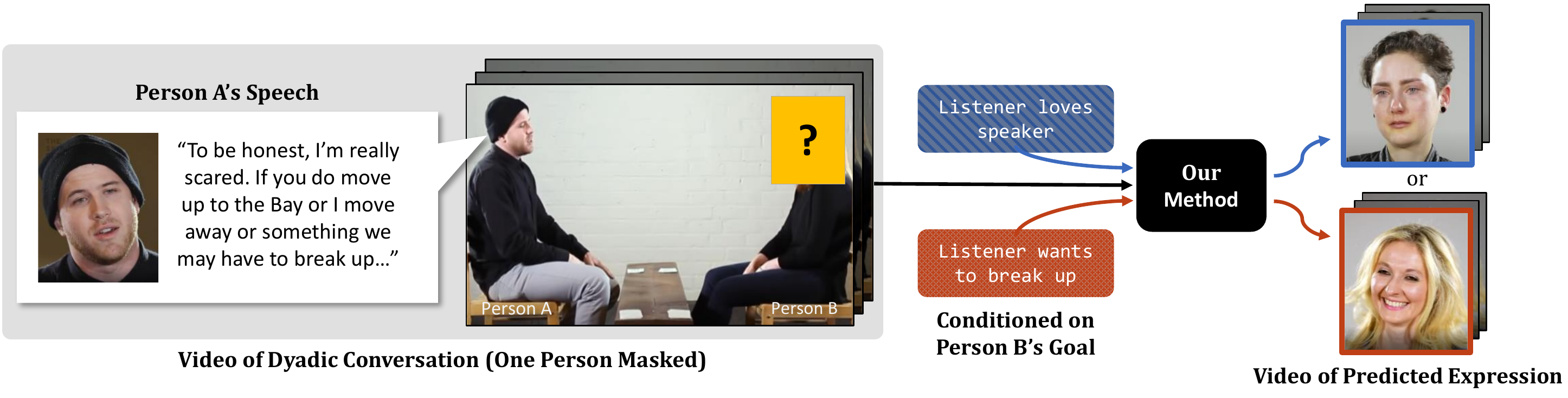}
\vspace{-2em}
\captionof{figure}{We introduce a method for modeling facial expressions during dyadic conversation between two people (Person A and Person B). Given the speech transcript of a speaker (left), and conditioned on the goal of a listener (middle), our method predicts the possible facial expressions of the listener (right). For example, if the listener in the above scenario wants to maintain their relationship with the speaker, they would likely have a serious expression. Our method is able to model this conversation in order to predict this expression.}
\label{fig:teaser}
\end{minipage}}
\end{picture}%

%% file: sec/1_abstract.tex
\begin{abstract}
We introduce a video framework for modeling the association between verbal and non-verbal communication during dyadic conversation. Given the input speech of a speaker, our approach retrieves a video of a listener, who has facial expressions that would be socially appropriate given the context. Our approach further allows the listener to be conditioned on their own goals, personalities, or backgrounds. Our approach models conversations through a composition of large language models and vision-language models, creating internal representations that are interpretable and controllable. To study multimodal communication, we propose a new video dataset of unscripted conversations covering diverse topics and demographics. Experiments and visualizations show our approach is able to output listeners that are significantly more socially appropriate than baselines. However, many challenges remain, and we release our dataset publicly to spur further progress. See our website for more details, including video results, data, and code: \href{http://realtalk.cs.columbia.edu}{\textbf{\textcolor{purple}{\texttt{realtalk.cs.columbia.edu}}}}.
\end{abstract}

%% file: sec/2_intro.tex
\section{Introduction}
\label{sec:intro}

If someone told you that they loved you, how would you react? Your facial expressions would vary wildly depending on whether it was your partner speaking or a stranger. A smile communicates a very different non-verbal message from a frown, and your choice of facial expression in this situation depends on your own background and goals. In computer vision, modeling non-verbal communication and building artificial social intelligence remains an important open challenge, with many applications in healthcare, assistive technology, and augmented reality. 

In this paper, we introduce a framework to model the relationship between verbal and non-verbal communication during dyadic conversation. Given the speech of one person, and the background of a second person who is listening, our system is able to produce a video of the facial expressions that the second person would likely display. Figure \ref{fig:teaser} shows one example result where the man discusses ending a personal relationship with another person, and our system is able to find possible listeners, which differ depending on the goal of the listener. 

\enlargethispage{-6.6cm} 

The computer vision community is actively exploring methods for modeling non-verbal communication ~\cite{jonell2020let, Joo_2019_CVPR, feng2017learn2smile, Ng_2021_CVPR, Fan_2021_CVPR}, including for the task of generating a listener from a speaker~\cite{ng_learning_listen, jonell2020let}. One of the key challenges is capturing the semantic knowledge that people have about the many topics that come up during conversation. However, most prior methods~\cite{ng_learning_listen, feng2017learn2smile, huang2017dyadgan, jonell2020let} have instead focused on modeling low- or mid-level features about the conversation, such as the tone and prosody of voice. Due to the lack of scalable training data, most methods have not been able to explicitly model the rich internal goals and backgrounds of real people, which is not directly observable in videos. 

We introduce a framework that composes large language models (LLMs) with vision-language models (VLMs) to associate listeners with speakers in videos. LLMs, such as GPT-3 \cite{brown_language_2020}, can be thought of implicit knowledge bases, condensing the information on the Internet into a generative language model. We capitalize on the few-shot learning capabilities of these models to identify patterns between verbal and non-verbal communication. Given the transcript of a speaker, our method prompts the LLM to generate the visual attributes of a listener, which we can consequently combine with a VLM such as CLIP\cite{radford_learning_2021} to retrieve a corresponding video of the facial expressions. Prompting provides control that we would otherwise not have, allowing us to condition the listener on different backgrounds or goals.

On a new dataset that we will release, our approach significantly outperforms prior work at predicting appropriate listeners to a speaker during conversation. Our dataset, \textit{The RealTalk Dataset}, contains videos of natural, unscripted conversations between two people, who come from diverse demographic backgrounds and discuss a wide range of topics and emotions. Through both automatic perceptual metrics as well as a human study, our results show the pivotal role of language models in building vision systems capable of reasoning about dyadic human social interactions. We encourage readers to see the supplemental material, which highlights qualitative videos of our results.

The main contribution of this paper is to introduce a new framework and dataset for modeling both verbal and non-verbal dyadic conversation, and the remainder of this paper will describe this contribution in detail. In section~\ref{sec:related}, we first summarize related work in human activity analysis, including in non-verbal communication. In section~\ref{sec:method}, we present our framework. In section~\ref{sec:dataset}, we describe our new dataset in detail, and in section~\ref{sec:experiments}, we use this dataset to compare our framework against prior methods as well as discuss limitations. We release all code, models, and data.

%% file: sec/3_related_framefig.tex
\section{Related Work}
\label{sec:related}
\textbf{Interactive Conversational Avatars}:
Prior works on conversational avatars often use manual hand-crafted features of lab-captured motion data \cite{bohus2010facilitating, cassell1994animated, gratch2006virtual, huang2011virtual, sonlu2021conversational, wang2011towards}. These approaches are limited in capturing the true complexity of facial expressions during interaction, and are difficult to generalize to in-the-wild data; there has thus been a recent push for data-driven approaches \cite{feng2017learn2smile, nojavanasghari2018interactive, jonell2020let, greenwood2017predicting,huang2017dyadgan, ng_learning_listen}. The state-of-the-art work, Learning2Listen\cite{ng_learning_listen}, autoregressively generates (goal-unconditoned) listener face models given audio and face features from a speaker. \cite{ng_learning_listen} and similar existing works focus evaluation on capturing low- or mid-level conversation features such as speaker-listener synchrony; they do not explicitly model the high-level semantics of the conversation, like what the speaker is communicating. We focus on modeling these high-level features, and their interaction with the goal of the listener.

\textbf{Reasoning with Language Models}: Large language models (LLMs) \cite{brown_language_2020, chowdhery2022palm, zhang2022opt}, through their internet-scale pretraining, capture implicit knowledge \cite{petroni-etal-2019-language} that can be used for a number of downstream tasks \cite{talmor2020leap, kojima2022large, creswell2022faithful}. A rich line of work in NLP has studied LLMs for social, behavioral, and cultural reasoning \cite{sap_social_2019, sap-etal-2020-commonsense, fung2022normsage, ziems-etal-2022-moral}. Performance in such multi-step reasoning tasks is known to be sensitive to how the model is prompted \cite{brown_language_2020}; \cite{wei_chain_2022} observe that \textit{chain of thought} prompting, in which a model is asked to produce intermediate outputs, are critical to strong performance. We extend these efforts and use GPT-3's implicit knowledge about human behavior to reason over social goals and non-verbal behavior in dyadic conversations.

\textbf{Vision-Language Recognition}: 
The recent paradigm of vision-language pretraining\cite{jia_scaling_2021, radford_learning_2021, singh2022flava, yuan2021florence, yu2022coca}, in which models are trained on large corpora of image-text pairs, has enabled vision to grow past the fixed-category paradigm. Models such as CLIP \cite{radford_learning_2021} and ALIGN \cite{jia_scaling_2021} learn a joint representation over images and text via a contrastive loss that pulls corresponding image-text pairs together in representation space, while pushing non-corresponding pairs apart. These models thus provide a notion of similarity across modalities, enabling recognition with flexible, freely-specified natural language. Previous work has leveraged this capability by capturing world knowledge as text to improve performance on vision tasks \cite{shen_k-lite_2022, zeng_socratic_2022, yang_empirical_2021, menon_visual_2022}. For example, \cite{yang_empirical_2021} uses GPT-3 to improve visual question-answering by prompting GPT-3 to reason about the textual summaries of images. \cite{menon_visual_2022} uses GPT-3 to obtain semantic descriptors of object categories, then compare query images to these descriptors using CLIP. This ``classification via description" improves accuracy and enables interpretable decision-making. While applied in the context of very different tasks from ours, such work highlights the potential of reasoning with language models and connecting the results back to the visual world, a motivation our work shares. 

\begin{figure*}
    \centering
    \includegraphics[width=\linewidth]{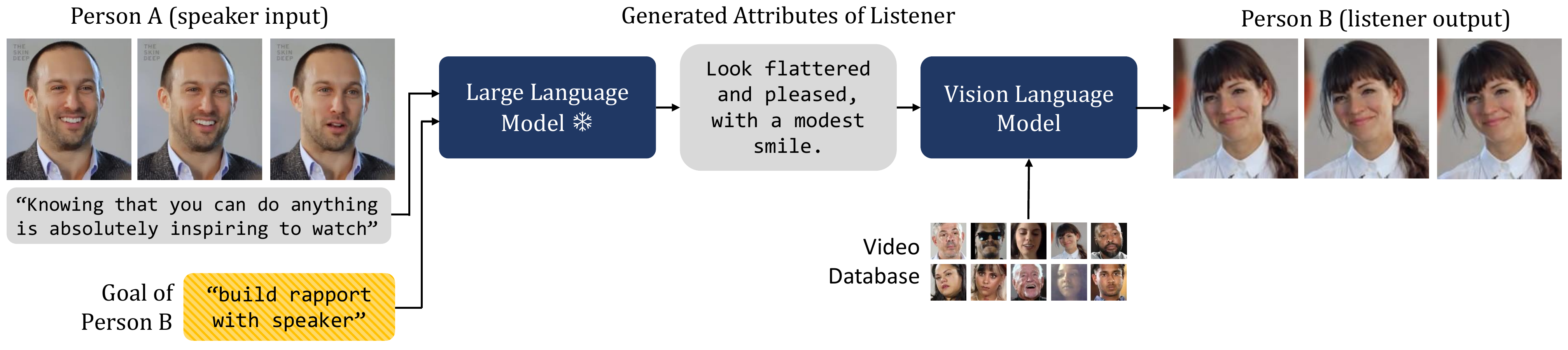}
    \vspace{-2em}
    \caption{\textbf{Overview of the Framework:} Given the transcript of one person (left), our method learns to retrieve possible videos of listeners (right). Since the emotional display of a person depends on their goals and intentions, our approach is also conditioned on the goals of the listener (yellow box). Solving this problem is challenging because the topics of conversations can be diverse. Our framework first uses a language model as an implicit knowledge base in order to generate descriptive attributes of the ideal listener (middle), then uses a vision-language model to retrieve possible listeners that match the descriptors.\vspace{-1em}}
    \label{fig:framework}
\end{figure*}

\textbf{Adapting Large-Scale Models}: 
Although large-scale vision-language models demonstrate impressive capabilities on a range of tasks when used in a purely zero-shot setting, there remain scenarios when some adaptation of the model is desired. The most obvious adaption method is finetuning, explored extensively by \cite{wortsman_robust_2022}, but finetuning requires substantial resources as all parameters of the large model must be updated. 
Similarly to the large language model setting, various methods have been developed to avoid full finetuning by instead learning a lightweight prompt. The prompt can be in textual space \cite{zhou_learning_2021, ju2022prompting}, where embeddings of prompt tokens are learned. For vision-language models, the prompt can also be in visual space. \cite{bahng_exploring_2022,jia_visual_2022} show two methods to learn visual prompts on a per-dataset basis, while \cite{menon_representational_2022} highlights prompting to guide representations towards particular tasks. All of the aforementioned works learn prompts from human-provided labels. In our work, we will explore adaptation when labels are unavailable.

%% file: sec/4_method.tex
\section{Retrieval by Describing Emotions}
\label{sec:method}

We now present our approach for modeling the association between non-verbal and verbal communication. We focus on dyadic conversation consisting of two people engaging in natural, unscripted conversation.

\subsection{Framework}

Given an input video of a speaker with transcript $x$, and a listener with the social goal of $g$, our aim is to output a video of a person with the facial expressions  and motions that best serve the specified goal. We represent the social goal $g$ as a natural language phrase, \eg ``build rapport.''

We formalize the problem as a retrieval task. Let $\mathcal{Y}$ be a fixed set of possible listener videos.   We estimate the best listener video through the mapping:
\begin{align}
\hat{y} = \argmax_{y \in \mathcal{Y}} \; \phi\left(c, y\right) \quad \textrm{where} \quad c = f(x,g)\quad 
\end{align}
such that $f(x,g)$ estimates semantic features that the listener should have to achieve $g$ while listening to $x$, and $\phi$ is a learned metric that scores how well $c$ matches $y$. 

This problem is challenging because we need to model potentially sophisticated topics in both the dialog $x$ and the social goal $g$, while also modeling the relationship between them and the visual modality. We will model $\phi$ through a composition of large language models (LLMs) and vision-language models (VLMs). By prompting an LLM to describe the facial expressions of the ideal listener conditioned on the goal, we can ground the descriptive attributes into the visual world with a VLM. Due to the amount of text that LLMs have encountered on the Internet, we leverage them as a knowledge base of social behavior. Figure \ref{fig:framework} shows an overview of this framework, which we now detail next.

\subsection{Reasoning on Listener's Goals}

Given a transcript $x$ of words spoken by a speaker and the listener goal expressed in text form $g$, we ask GPT-3 to describe approximate listener expressions, which we use to form $c = f(x,g)$. We accomplish this by providing GPT-3 with the following prompt:
\begin{lstlisting}[breakatwhitespace=true]
Alex says to Sam: {x}
Q: What is Alex communicating to Sam? Given that Sam wants to {g}, describe Sam's facial expressions in visual detail.
A: 
\end{lstlisting}
where  \lstinline|{x}| and \lstinline|{g}| are substituted with the transcript and goal respectively. We intentionally choose gender-neutral names in the prompt to mitigate bias. The response to this query provides a description of the desired listener's features in natural language. For example, if a speaker says ``My cat passed away yesterday..." and the listener wants to behave in a ``socially appropriate" way, the response from GPT-3 would be:
\begin{lstlisting}[breakatwhitespace=true]
Alex says to Sam: "My cat passed away yesterday"
Q: What is Alex communicating to Sam? Given that Sam wants to behave in a socially appropriate way, describe Sam's facial expressions in visual detail.
A: <@\texttt{\textcolor{darkgreen}{Given that Sam wants to behave in a socially appropriate way, Sam should look sad and sympathetic.}}@>
\end{lstlisting}
We directly use the GPT-3 response, e.g. ``Sam should look sad and sympathetic," as the descriptive, semantic features $c$ for our desired listener, which will be used to retrieve a listener video. We also find that chain-of-thought reasoning \cite{wei_chain_2022} and few-shot prompting improve the generated descriptive features by the language model. Further implementation details can be found in the supplementary.

\subsection{Grounding Attributes in Video}

Once we have the desired listener facial expression in the form of natural language text, our next task is to ground the semantic attributes $c$ into visual content of listener videos $\mathcal{Y}$ in order to retrieve the video that best aligns with it. In short video clips ($\leq 15$ seconds), the facial expression of the listeners is relatively stable and does not change much over time. We therefore compute the similarity with a large-scale vision-language model (VLM):
\begin{align}
\phi(c,y) = \psi_\textrm{txt}(c) \cdot \sum_{t \in \mathcal{N}(y)} \frac{\psi_\textrm{img}(y_t)}{|\mathcal{N}(y)|}
\label{eqn:visual_sim}
\end{align}
where $y_t$ represents a frame in the video $y$, and $\psi_\textrm{txt}$ and $\psi_\textrm{img}$ are the text and visual encoders of a VLM respectively. We use $\cdot$ to indicate the dot product. In this work, we implement $\psi$ with CLIP ViT-B/32~\cite{radford_learning_2021}. We compute the mean over the set of key frames $\mathcal{N}(y)$ that have the most pronounced facial expressions, which we find automatically through the norm on the features extracted by EMOCA~\cite{danvevcek2022emoca}; we refer readers to the supplementary material for full details.

\begin{figure}[t]
    \centering
    \includegraphics[width=\linewidth]{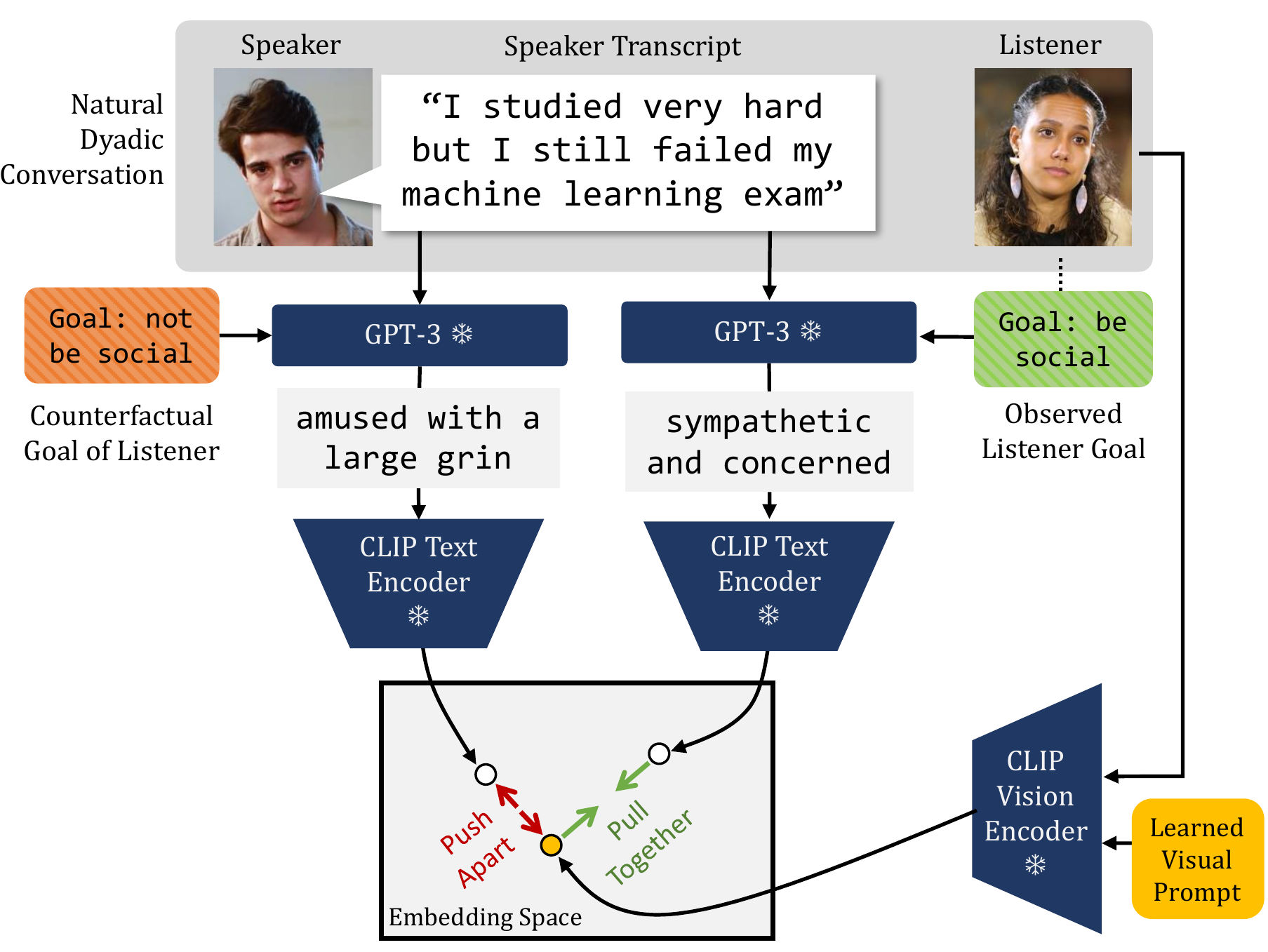}
    \caption{\textbf{Adapting CLIP:} We adapt CLIP to our data distribution by training a visual prompt with a contrastive loss that pulls ground-truth listener towards the GPT-3 predicted socially-correct response, and away from the counterfactual, socially-incorrect response. However, while typical videos contain positive pairs of speakers/listeners, most videos do not contain negative pairs. We propose to use GPT-3 to generate such negative examples.}
    \label{fig:contrastive}
\end{figure}

\subsection{Refining the Visual Representation}
We found that, although state-of-the-art VLMs such as CLIP are trained on Internet-scale image-text paired data, they are empirically poor at aligning human expressions in images with their corresponding semantic attributes. The standard solution is to adapt the VLM on a labeled dataset for the target task distribution (\eg via finetuning), which has been shown to significantly improve performance on most other tasks. Such adaptation uses a contrastive loss, which requires positive and negative ground-truth examples. However, in our setting, most videos in the dataset show successful conversations; we do not have natural access to negative pairs of conversations. 

However, since language models are generative models, we do have access to a mechanism to synthesize scenarios in text. We propose to leverage this capability to automatically construct training examples and therefore finetune VLMs on our task. We assume that, in our dataset, most conversations are successful and people generally display socially appropriate emotions given the conversation. We are able to use these examples as positive examples. To generate  negative examples, we prompt GPT-3 to describe the non-verbal features that would derail the conversation and be socially inappropriate. For example, if someone is discussing that they failed a machine learning exam, the language model can predict that amusement is the wrong emotion to display.

Since full finetuning of large models can be computationally prohibitive, our approach refines the visual representation by optimizing a visual prompt $\delta$ that uses a small number of parameters while keeping the parameters of $\psi$ fixed. Over a training set of dyadic conversation videos, we optimize with gradient descent:
\begin{align}\min_\delta  \;  \mathbb{E}_{(x,y)}\left[\log
\frac{\alpha(g)}{\alpha(g) +\alpha(\neg g)}
\right] \label{eqn:finetune}
\end{align}
where $\alpha(g)$ is the distance between the conversation and the social goal $g$. We use $\neg g$ to represent the natural language sentence that is the opposite of $g$, and we assume $g$ is ``be social'' and $\neg g$ is ``not be social.'' We compute the distance:
\begin{align}
 \alpha(g) = \exp\{ \lVert \psi_\textrm{img}\left(y_t \cup \delta\right) - \psi_\textrm{txt}\left(f(x,g)\right) \rVert_2 \}
\end{align}
where the visual prompt $\delta$ is concatenated to the input frame $y_t$. By optimizing Equation \ref{eqn:finetune} (equivalent to optimizing an InfoNCE loss~\cite{oord_representation_2019}), the language model $f$ will automatically construct negative examples, allowing the visual representation $\psi_\textrm{img}$ to be refined for this task, even when there is limited training data available.  We use the refined $\psi_\textrm{img}\left(y_t \cup \delta\right)$ to form $\psi_\textrm{img}$ during evaluation. Figure \ref{fig:contrastive} shows an overview of this refinement setup.

%% file: sec/5_dataset.tex
\section{The RealTalk Dataset}
\label{sec:dataset}
\begin{figure}
    \centering
    \includegraphics[width=\linewidth]{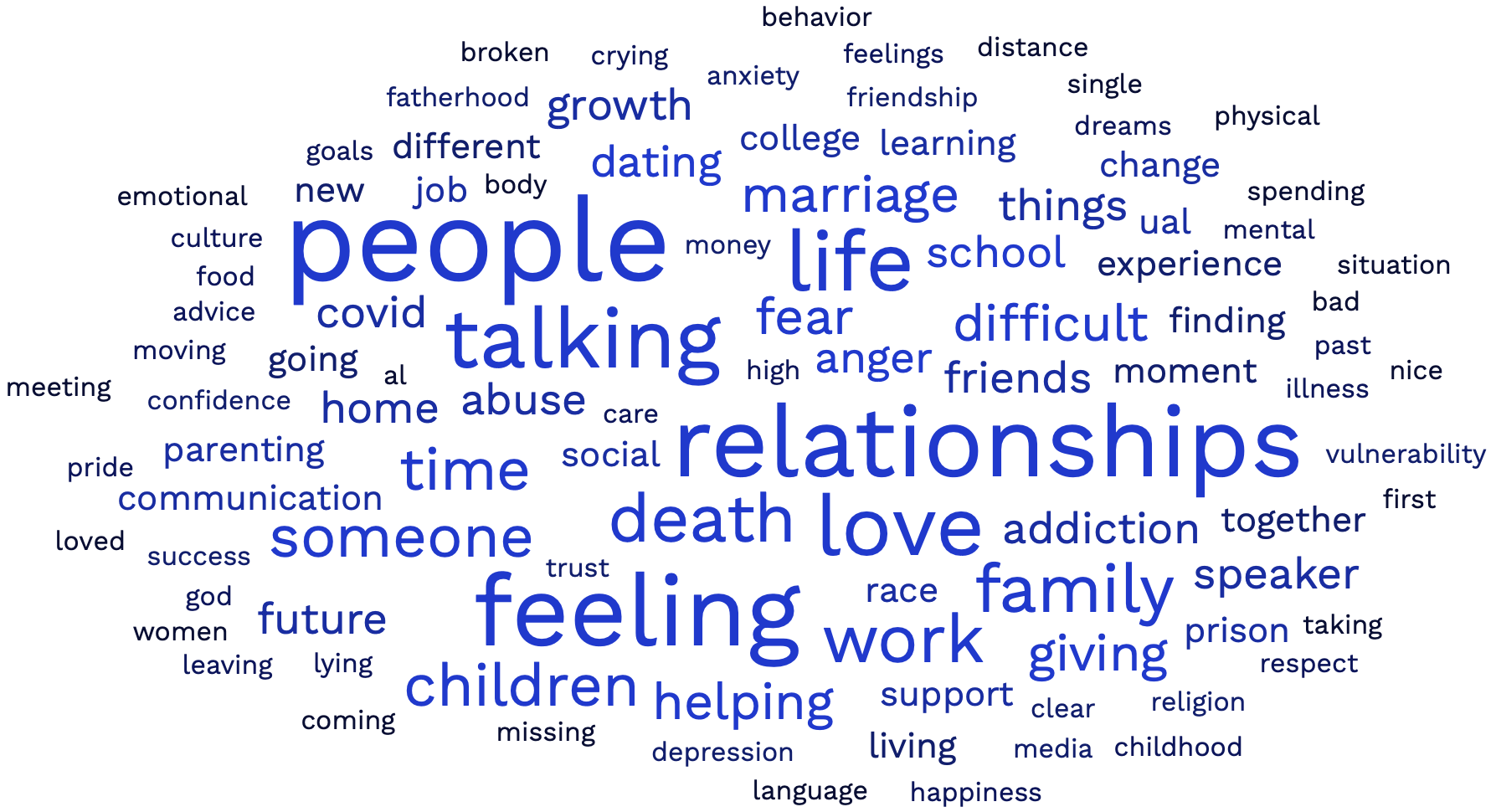}
    \vspace{-0.5em}
    \caption{\textbf{RealTalk Topics:} Our dataset covers diverse conversation topics. We used GPT-3 to estimate the topic of the conversation, and we show a ``word cloud'' of the topics where the font size indicates frequency.}
    \label{fig:wordcloud}
    \vspace{-1em}
\end{figure}

\begin{figure}
    \centering
    \includegraphics[width=0.86\linewidth]{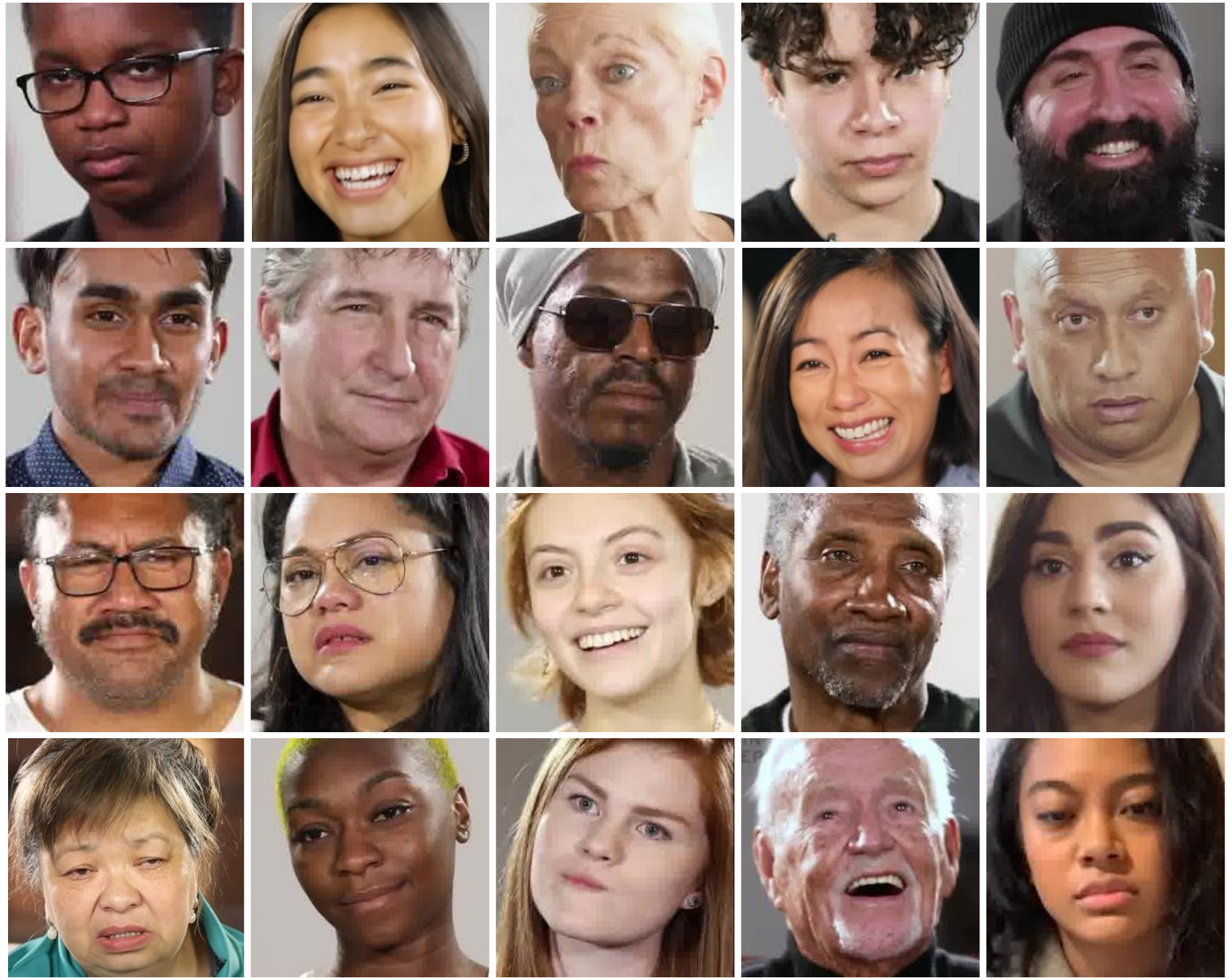}
    \vspace{-0.5em}
    \caption{\textbf{RealTalk Faces:} The dataset contains people from diverse demographics and includes a range of facial expressions. Since the conversations are unscripted, the expressions are natural and not exaggerated, making the dataset a challenging benchmark.}
    \label{fig:datasetviz}
     \vspace{-1em}
\end{figure}

We present the RealTalk dataset for studying human dyadic non-verbal and verbal interaction. The dataset consists of 692 in-the-wild videos from \textit{The Skin Deep}, a popular YouTube channel capturing long-form, unscripted personal conversations between diverse individuals about different facets of the human experience. As shown in Figure~\ref{fig:wordcloud}, conversations in our dataset deal with topics such as family, dreams, relationships, illness, mental health, and many more. Due to the nature of these topics, our dataset organically captures a wide gamut of emotions and expressions. These expressions are readily visible; a key advantage of our dataset is that both individuals in each conversation directly face the camera, against a blank background. We show a representative subset in Figure~\ref{fig:datasetviz}. 

On average, each video in our dataset is slightly over 10 minutes long, adding up to a total of 115 hours of raw data. We will publicly release the dataset, along with pre-computed ASR transcripts, visual embeddings, and speaker annotations. We believe that this dataset will benefit the development and evaluation of computer vision methods that analyze human emotion and interaction.

\subsection{Data Collection and Annotation}
We collect our videos from YouTube at 25 frames per second and 1280x720 resolution. We run an off-the-shelf face detection model~\cite{bulat2017far}, and keep only frames where both individuals of the conversation are fully visible. We found this step was sufficient to remove video artifacts. We generate transcripts for each video with automatic speech recognition (ASR)~\cite{radford2022robust}, and we annotate each frame with the speaking individual(s) using active speaker detection~\cite{tao2021someone}.

\subsection{Benchmark}
We use our data as a benchmark to evaluate our method on goal-conditioned listener retrieval. Using our active speaker annotations, we isolate clips from our dataset where only a single person is speaking, and mask the other person as the listener to be predicted. Since it is difficult, even for humans, to gauge the emotional content of a conversation without sufficient context, we construct long-form (384 frames, or 15.36 seconds) clips that capture complete thoughts. In total, this process yielded 1896 clips. We use 1489 clips to train our model's visual prompt, and we evaluate on the remaining 407. We use the full set of 1896 clips as the retrieval databank.

%% file: sec/6_experiments.tex
\section{Experiments}
\label{sec:experiments}

\begin{figure*}
    \centering
    \includegraphics[width=\linewidth]{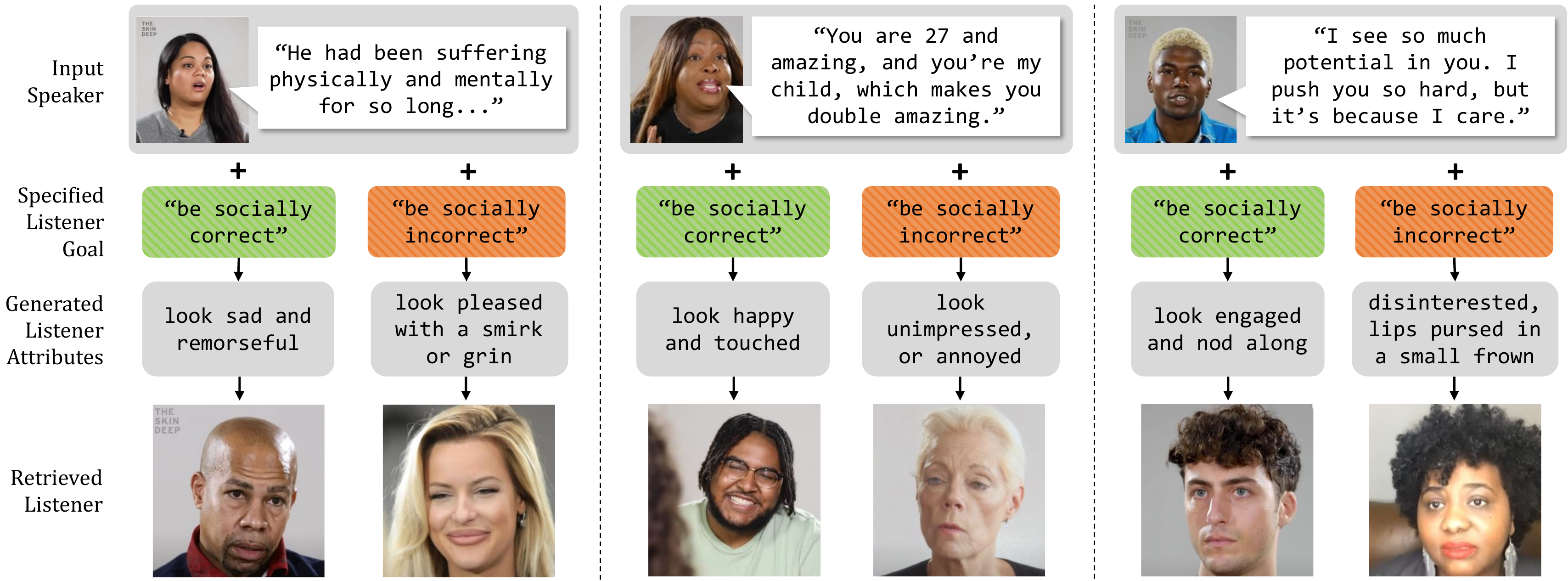}
    \vspace{-2em}
    \caption{\textbf{Qualitative Examples:} We highlight the output of our method on three representative samples from our dataset. For each speaker, we retrieve videos of two listeners, one whose goal is to act socially correctly, and one adversarial listener whose goal is to act socially inappropriately. Our outputs reflect this dichotomy in goal: conditioning on the listener's goal drastically changes the listener attributes generated by GPT-3, which is reflected in the visual expression of our final retrieved listeners.\vspace{-1em}}
    \label{fig:qualitative_examples}
\end{figure*}

\begin{figure*}
    \centering
    \includegraphics[width=\linewidth]{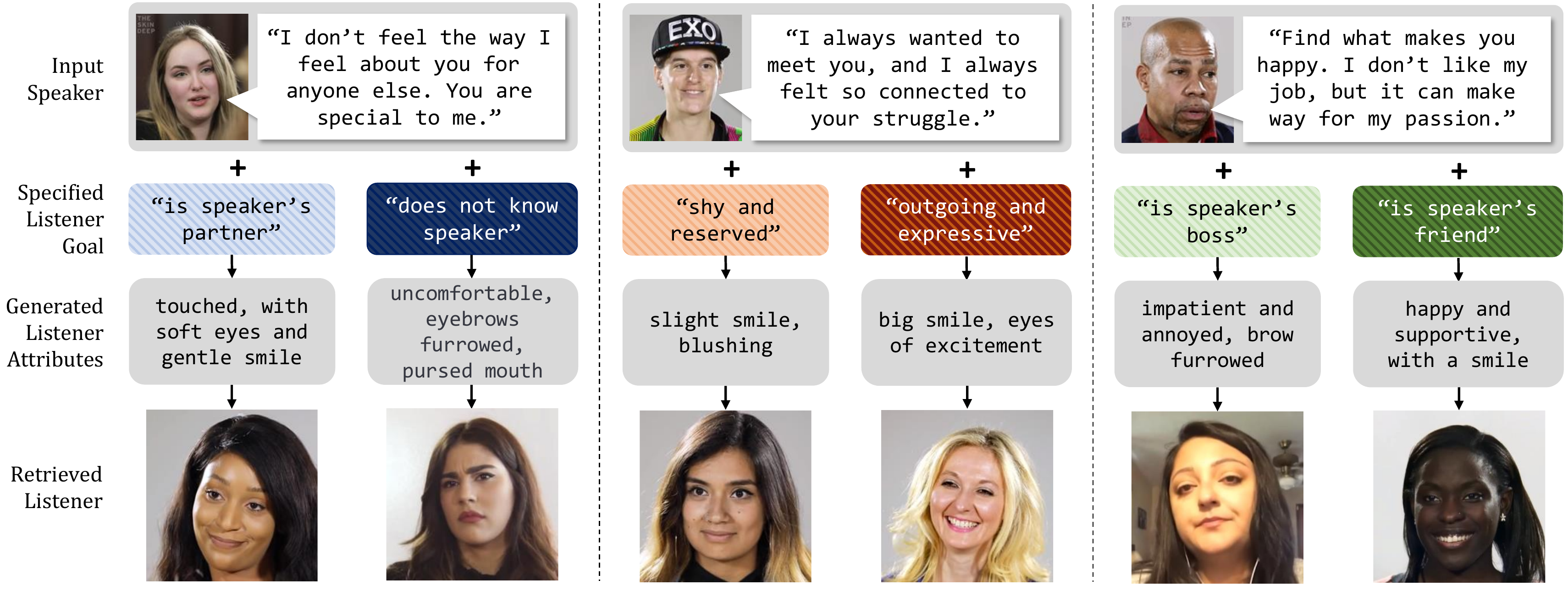}
    \vspace{-2em}
    \caption{\textbf{Open-World Goals:} Our framework can often generalize to open-world listener goals, such as the listener's personality traits and their relationship with the speaker. In these cases, GPT-3 correctly infers the implied consequences, yielding correct listener retrievals.}
    \label{fig:extra_goals}
\end{figure*}

The goal of our experiments is to evaluate the overall social appropriateness of our predicted listeners, and to study the degree to which our predicted listeners respect the input goals. While our method naturally scales to handle arbitrary listener goals (Figure~\ref{fig:extra_goals}), we focus our evaluations on two dichotomous goals that are broadly applicable across social scenarios: (1) behave socially appropriately, and (2) behave socially inappropriately. We present a representative qualitative selection of listeners produced with these goals in Figure~\ref{fig:qualitative_examples}, and detail quantitative experiments below.

\subsection{Baselines}
We compare our predicted ``socially correct" and ``socially incorrect" listeners to the closest related work, Learning2Listen, along with natural baselines and ablations.

\textbf{Learning2Listen (L2L):} The SOTA method for speaker-aware 3D avatar generation is Learning2Listen \cite{ng_learning_listen}, which we retrain on our data using the official implementation. Given speech audio and face model of a speaker, L2L generates the corresponding (goal unconditional) listener as a sequence of parameters to a 3D face model. Since we are interested in listener video retrieval, we use L2L's generated output to perform retrieval by finding the nearest neighbor in face model parameter space.\footnote{The original paper \cite{ng_learning_listen} also has a method to generate photorealistic faces. However, it requires proprietary subroutines that could not be shared with us to make comparisons.}

\textbf{Untuned CLIP:} This baseline uses our framework, except with our contrastive learning procedure ablated. Listener video retrieval is performed with CLIP frozen to the public pretrained weights, with no additional task-specific adaptation. Comparison against this baseline measures the gain provided by our learned visual prompt.

\textbf{Uniform Frame Sampling:} This baseline also uses our framework, except the similarity metric between a listener video to text attributes is computed via uniformly-spaced frames from the video, as opposed to importance-sampled key frames. In the notation of Equation~\ref{eqn:visual_sim}, where we detail this computation, the set of key frames $\mathcal{N}(y)$ for a given video $y$ is replaced with a set $\mathcal{N'}(y)$ consisting of uniformly-spaced frames from $y$.

\textbf{Random Chance:} We compare against the naive prior of randomly selecting a listener from the retrieval set. Intuitively, comparison against chance allows us to compare against listeners that move and behave realistically, but do not take the speaker's words or voice into context.

\subsection{Automatic Quantitative Metrics}
\begin{figure}
    \centering
    \includegraphics[width=\linewidth]{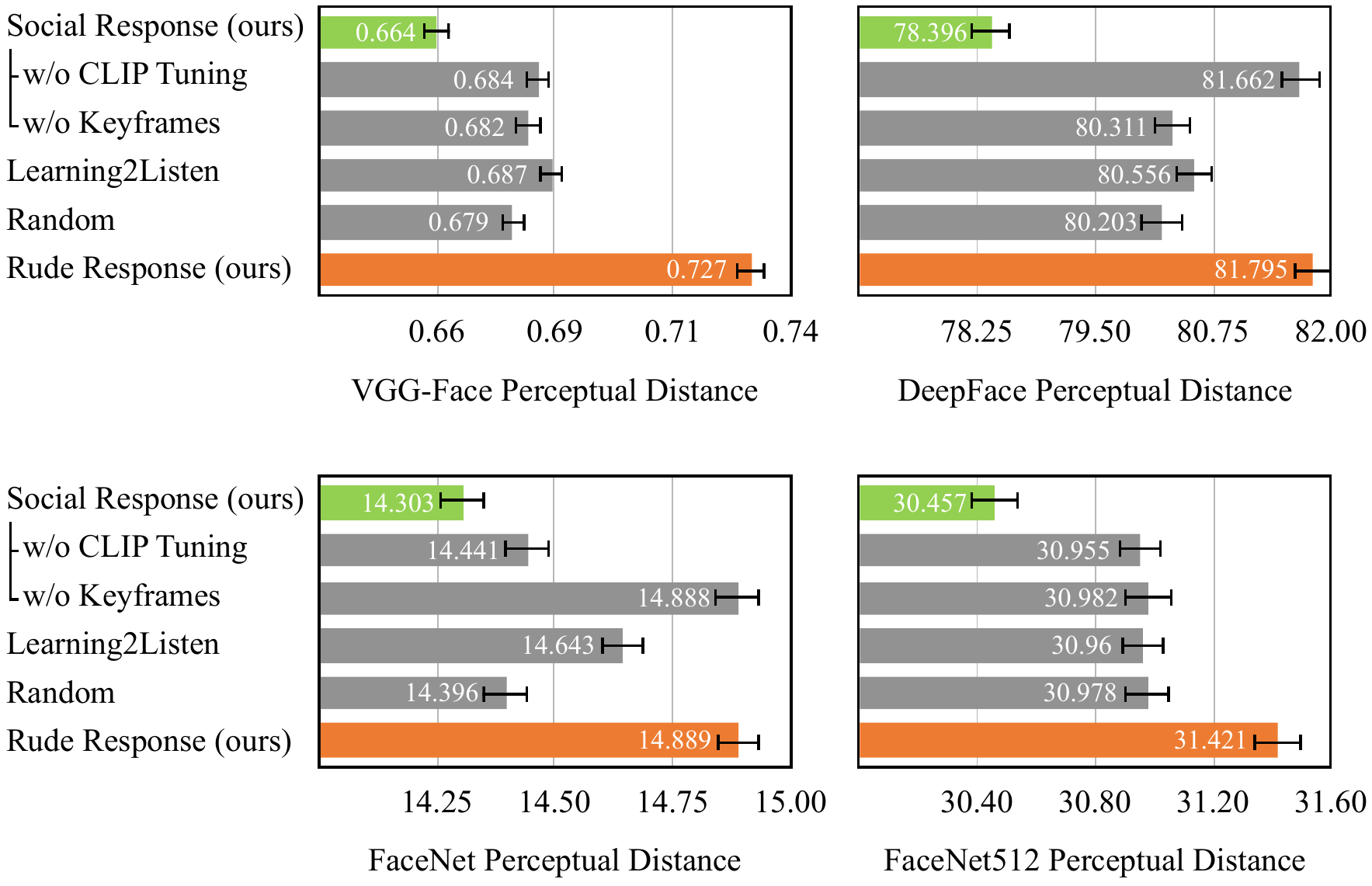}
    \vspace{-2em}
    \caption{\textbf{Perceptual loss} between predicted listeners and the ground-truth listener, computed as L2 distance in the representation space of several pretrained face models (lower is better). Our predicted socially correct listener yields the lowest perceptual loss across all methods. Error bars represent 95\% confidence intervals.\vspace{-1em}}
    \label{fig:perceptualloss}
\end{figure}

In our dataset, the people almost always engage in socially correct conversation, \emph{i.e.} the participants are rarely rude to each other. Therefore, we measure the quality of our predicted listener by how well they match the actual listener in the video.

\textbf{Perceptual Loss:} Similar to the Inception score~\cite{salimans2016improved, borji2022pros},
we quantify the perceptual similarity between our predicted listener and the actual listener by using the distances in the feature  spaces of several state-of-the-art deep networks pretrained on face images. Given a listener prediction, we compute a perceptual loss as the distance in representation space between the predicted listener and the ground-truth listener embeddings. Specifically, we use VGG-Face, FaceNet, FaceNet-512, and DeepFace~\cite{parkhi2015deep, schroff2015facenet, taigman2014deepface, serengil2020lightface}.

We report results in Figure~\ref{fig:perceptualloss}, which show that our retrieved ``socially correct" listeners are consistently and statistically significantly closer to the ground-truth listener in perceptual distance than listeners produced by any other baseline. Contrastingly, our retrieved ``socially incorrect" listeners are consistently \textit{furthest} in perceptual distance. Given that ground-truth listeners often behave socially correctly, this result suggests that the ``socially incorrect" listeners predicted by our framework do respect their specified goal, corroborating the qualitative results in Figure~\ref{fig:qualitative_examples}. Ablating either our CLIP adaption procedure or our keyframe sampling procedure reduced our performance to random guessing or worse, highlighting the importance of each method component. In particular, the performance gain conferred by adapting CLIP with GPT-mined negatives provides promising evidence for the potential of using large language models to counterfactually generate pseudo-labels for unobserved phenomenon.

\begin{figure}
    \centering
    \includegraphics[width=\linewidth]{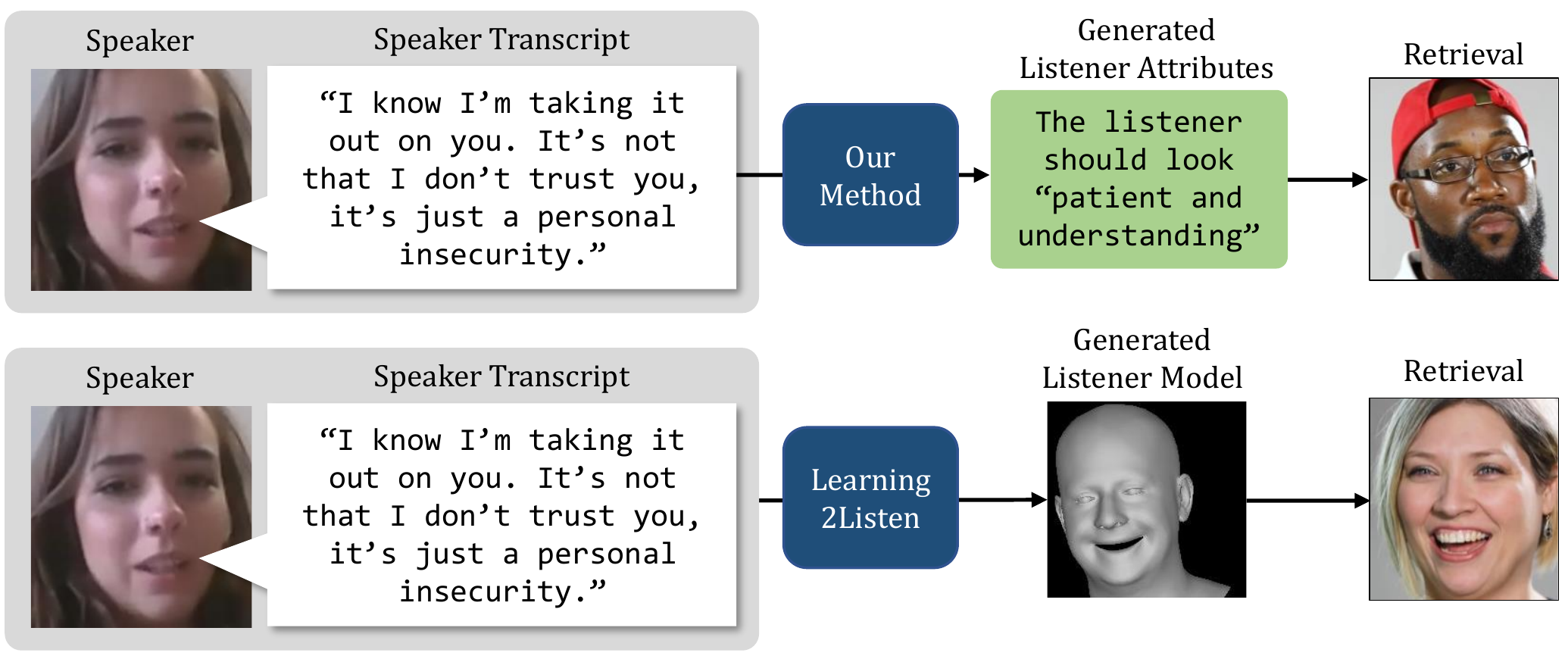}
    \vspace{-2em}
    \caption{\textbf{Method Comparison}: Learning2Listen does not take speaker ASR as input, and will sometimes ignore the semantic meaning of the speaker's message, leading to a socially inappropriate listener. Our framework, which reasons about language by construction, is able to handle such cases correctly.\vspace{-1em}}
    \label{fig:comparel2l}
\end{figure}

Intriguingly, listeners produced by Learning2Listen were not significantly different in perceptual quality from random listeners. To better understand this result, we visualize a Learning2Listen prediction in Figure~\ref{fig:comparel2l}. We found Learning2Listen---which does not take speaker ASR as input---often ignored the semantic meaning of the speaker's words, leading to smiling listeners in serious situations. Our method, which is constructed to reason about language, did not make such errors. Overall, our results suggest the importance of language in modeling social situations.

\textbf{Recall:}
We also evaluate our method by measuring its ability to retrieve the ground-truth listener in the top predicted listeners. We use Recall@$k$ (R@$k$), defined as the fraction of speakers in the evaluation set for whom the corresponding ground-truth listener is ranked among the top $k$ predictions, as well as the ground-truth listener's median retrieval rank (M. Rank) as our evaluation metrics. As Table~\ref{tab:recall_stat} shows, our method generally outperforms other baselines in retrieving ground-truth listeners. Absolute recall performance is relatively low across-the-board; this is reasonable, as there may be many listeners in the retrieval set who are perceptually similar to the ground-truth listener, and we do not model fine-grained listener motion. Performance is relatively similar regardless of whether or not we adapt CLIP. This is also reasonable, as our learning procedure only makes a lightweight adjustment to CLIP's representation space; this may beneficially affect which listeners are closest to a given semantic attribute (as suggested by the results in Figure~\ref{fig:perceptualloss}), but is unlikely to shift the overall structure across the retrieval videos.

\begin{table}[t]
\setlength{\tabcolsep}{2pt}
  \centering
  \begin{tabular}{lcccc}
    \toprule
    Method  &  R@500 $\uparrow$ & R@1000 $\uparrow$ & M. Rank $\downarrow$\\
    \midrule
    Social Response (ours)  & \textbf{0.31} & 0.56 & \textbf{834} \\
    -- w/o CLIP Tuning  & 0.29 & \textbf{0.60} & \textbf{830} \\
    -- w/o Keyframes  & \textbf{0.31} & 0.56 & 895 \\
    Learning2Listen  & 0.26 & 0.50 & 996 \\
    Random Chance  & 0.26 & 0.52 & 948 \\
    Rude Response (ours)  & 0.22 & 0.47 & 1061 \\
    \bottomrule
  \end{tabular}
  \vspace{-0.5em}
    \caption{\label{tab:recall_stat} \textbf{Comparison of retrieval performance} across baselines. We report median rank (M. Rank) and mean recall@$k$ over a 407 sample test set. The retrieval set has 1896 videos.}
\end{table}

\begin{table}[t]
\setlength{\tabcolsep}{2pt}
  \centering
  \begin{tabular}{lr}
    \toprule
    2-way comparison groups  &  \% times ours preferred\\
    \midrule
    Ours (social) vs. Learning2Listen & 71.7 $\pm$ 4.11 \\
    Ours (social) vs. $\neg{\text{Ours}}$ (rude)  &  77.5 $\pm$ 3.81\\
    Ours (social) vs. Ground-truth  &  20.0 $\pm$ 3.65\\
    \midrule
    Random Guessing & 50.0 $\pm$ 0.00\\
    \bottomrule
  \end{tabular}
  \vspace{-0.5em}
\caption{\textbf{Human Studies:} We show percentage of times our method's retrieved social listener is preferred over other baselines' retrieved videos by the study participants, when asked to choose the ``more socially-appropriate" listener among the two choices.\vspace{-1em}}
    \label{tab:human_study} 
\end{table}

\subsection{Human Studies}
To corroborate our quantitative results, we recruited college-age participants not involved in the research to judge the results. We use a two-alternative forced choice protocol. In each study, we presented participants with a series of speaker videos, along with two possible listeners per speaker: one listener retrieved by our method with the goal ``be socially correct;" and one listener produced by a comparison group. After watching the speaker side-by-side with each listener as many times as they wished, participants selected the listener they deemed as a more socially-appropriate reaction for the speaker. Since the social quality of a listener is most important at the emotionally-critical moments (\eg when the speaker compliments the listener, shares a regret, etc.) of a conversation, we manually curated such moments from our held-out test set by selecting speakers whose ground-truth listener displayed a non-neutral emotion. We randomly sampled 30 speakers from this set, and used this subset as the basis for our studies.

For each speaker, we compared our predicted ``socially correct" listeners to (1) our predicted ``socially incorrect" listeners, (2) Learning2Listen, and (3) the speaker's ground-truth listener. For each speaker, every listener comparison was made by 4 evaluators, yielding a sample size of $4 \times 30 = 120$ for each comparison group. We report the results in Table~\ref{tab:human_study}. Overall, our retrieved ``socially correct" listeners are preferred over both Learning2Listen and our ``socially incorrect" listeners at a statistically significant rate. Moreover, the degree of this preference is highest when comparing to our ``socially incorrect" listeners, which reflect the quantitative trends shown earlier in Figure~\ref{fig:perceptualloss}. A gap between our method and human performance remains; ground-truth listeners are consistently preferred over ours.

\subsection{Understanding Failure Cases}

\begin{figure}
    \vspace{-1em}
    \centering
     \begin{subfigure}[b]{\linewidth}
         \centering
         \includegraphics[width=\linewidth]{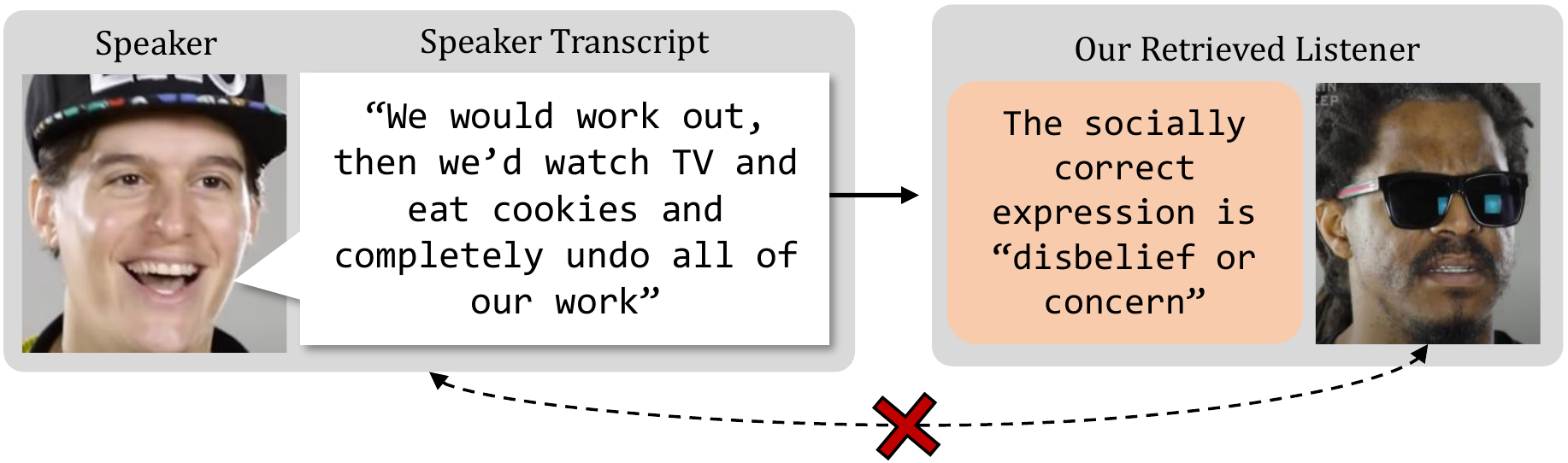}
         \caption{Failure due to lack of multi-modal speaker information. The speaker has a smile, which should modulate the meaning of their speech.}
         \label{fig:failure_multi}
     \end{subfigure}
    \vskip2mm
     \begin{subfigure}[b]{\linewidth}
         \centering
         \includegraphics[width=\linewidth]{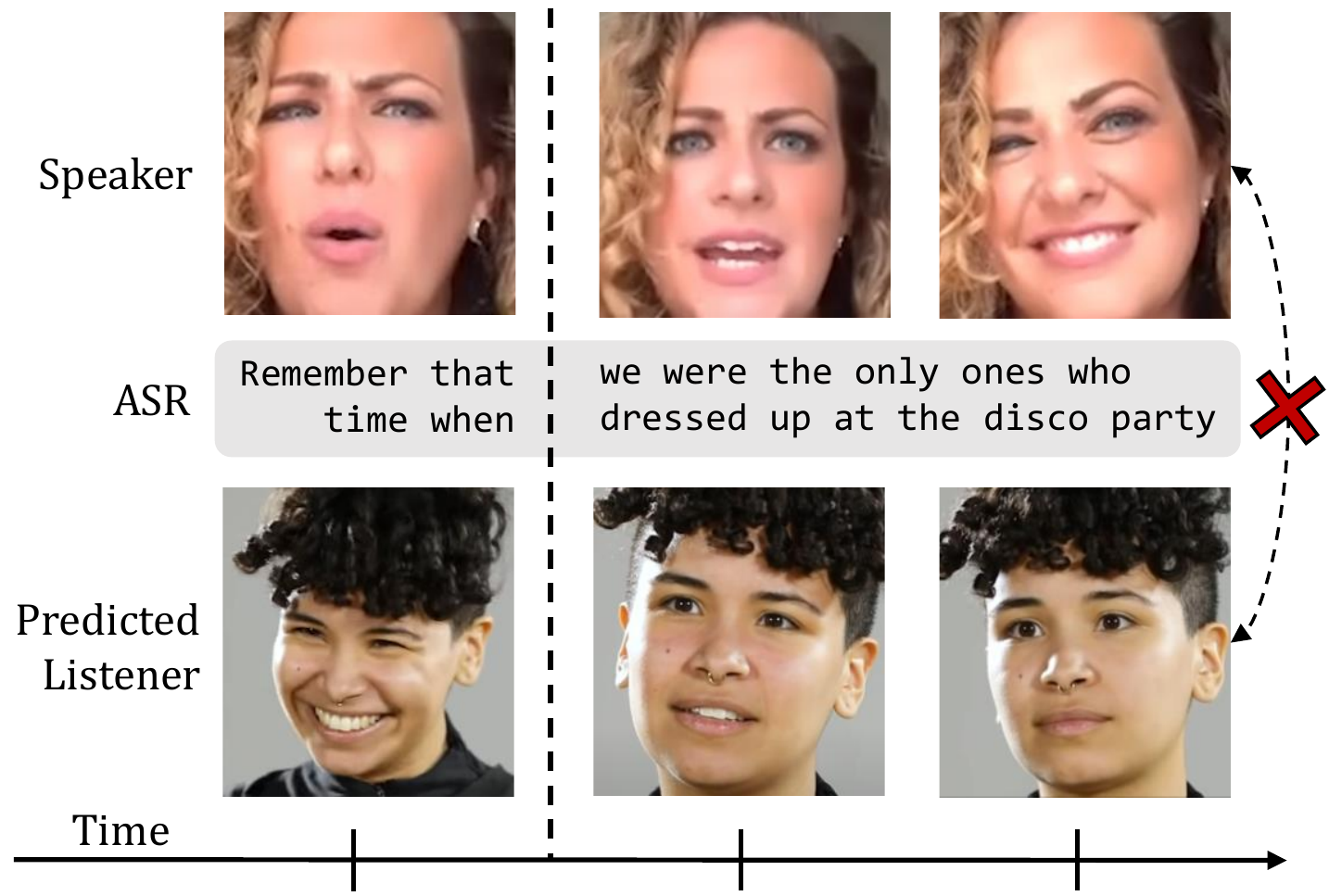}
         \caption{Failure due to temporal misalignment between speaker and listener. The retrieved listener smiles before the amusing story is conveyed.}
         \label{fig:failure_time}
     \end{subfigure}
    \vspace{-2em}
    \caption{\textbf{Failure Cases:} We qualitatively explore limitations of our method. In (a), our method fails to recognize that the speaker is making a self-deprecating joke, as we do not model the visual features of the speaker. This leads to an overly serious listener. In (b), our predicted listener laughs before the punchline of the story, as we do not model temporal dynamics in listener video.\vspace{-2em}}
    \label{fig:failures}
\end{figure}

We explore a few limitations of our method by presenting qualitative failures. Generally, we found that most failures were due to one of three reasons:

\textbf{1. Lack of multi-modal speaker context.} Given video of a speaker, our method only considers the language of what the speaker is saying. As our results demonstrate, this is often a fruitful simplification, but there are cases where the facial expressions and voice prosody of the speaker may alter the interpretation of their message. Consider the example in Figure~\ref{fig:failure_multi}. From the speaker's large smile, we can clearly tell that their speech is meant to be self-deprecating humor; the optimal reaction would be to laugh along. Instead, our method, relying on the speaker's negative words alone, predicts an overly serious listener.

\textbf{2. Temporal speaker-listener misalignment.} Our framework models listener videos as a set of key frames, and does not explicitly model any temporal dynamics between the speaker and listener. Thus, we get failures where the predicted listener has the correct expressions, but at the wrong time. For example, in Figure~\ref{fig:failure_time}, the listener laughs before the punchline of the speaker's story.

\textbf{3. Language model failures.}
Our method inputs speaker ASR into GPT-3 to reason about how the listener should react. However, GPT-3 occasionally misinterprets the speaker's words and consequently produces an incorrect response. We found this problem especially prevalent when the speaker used metaphors (which GPT-3 often interpreted literally), or when the speaker's words were vague. In such cases, we observed that GPT-3 would hallucinate incorrect details into the speaker's words (explored in \cite{maynez-etal-2020-faithfulness}). For example, in one sample the speaker solemnly says ``I just couldn't forgive myself,'' then GPT-3 hallucinates that the speaker is confessing to ``stealing the listener's bike," and thus incorrectly predicts that the listener should ``look angry or upset." As language models progress, we expect this failure mode to diminish; improvements in large models will directly improve our framework.

%% file: sec/7_conclude.tex


\section*{Acknowledgements}
\noindent  We thank Athena Tsu for her valuable assistance in creating visualizations, as well as D\'idac Sur\'is and Sruthi Sudhakar for their helpful feedback. This research is based on work partially supported by the DARPA CCU program under contract HR001122C0034 and the NSF NRI Award \#2132519. S.G.\ is supported by the Rabi Scholarship; S.M.\ is supported by the NSF GRFP fellowship. The views and conclusions contained herein are those of the authors and should not be interpreted as necessarily representing the official policies, either expressed or implied, of the sponsors.

%% file: sec/8_appendix.tex
\appendix
\newpage

\twocolumn[
\centering
\Large
\textbf{Appendix} \\
\vspace{1.0em}
]
\appendix

\section{Method Details}

\subsection{Video Keyframe Extraction}
We observe that human emotion is often sparse in video: on average, a person will have a blank expression, but the few key frames in which they do display emotion dominates our perceptual interpretation of their overall reaction. To automatically identify these key frames from an input listener video, we use EMOCA~\cite{danvevcek2022emoca}, a state-of-the-art facial expression extraction model, to regress a parametric 3D face model of the listener. The outputs include a disentangled expression parameter for each frame of the listener, in accordance to the FLAME linear 3D morphable face model~\cite{FLAME:SiggraphAsia2017}. FLAME's expression parameter space is constructed to be linear, giving the norm on the space a meaningful interpretation: the higher the norm of a listener's expression parameter, the more perceptually extreme their associated expression. Thus, we extract video keyframes as the frames with high associated expression parameter norm relative to other video frames. We identify these frames with the scipy.signal.find\_peaks function.

The above discussion is summaraized in Algorithm~\ref{algo:keyframe}.

\begin{algorithm}
\caption{Listener Video Keyframe Extraction}\label{algo:keyframe}
\begin{algorithmic}
\Require{A listener video $y$, an integer parameter $k$}
\Ensure{The set $\mathcal{N}(y)$ of the top $k$ most-expressive keyframes from $y$}

\ForAll {frame $y_i$ $\in$ video $y$}
        \State{Regress expression parameter $\phi_i$ of listener from frame $y_i$}
        \Comment{EMOCA}
\EndFor
\State{Compute peaks of the signal $||\phi_i||_2$ over frame indices $i$.} \Comment{Implemented with scipy.signal.find\_peaks}
\State Return the $k$ highest peaks as $\mathcal{N}(y)$.
\end{algorithmic}
\end{algorithm}

\subsection{Generating Listener Attributes with GPT-3}
\textbf{Few-shot Prompt Design:} We found that zero-shot GPT-3 was unable to effectively generate useful listener attributes, due to three main issues. First, the generated attributes were often overly broad and not visually descriptive. Second, the generated attributes often ignored or misinterpreted the speaker's ASR speech. Third, the generations varied in structure across samples, which is unsuitable for further automated processing.

\begin{figure*}
    \centering
    \includegraphics[width=\linewidth]{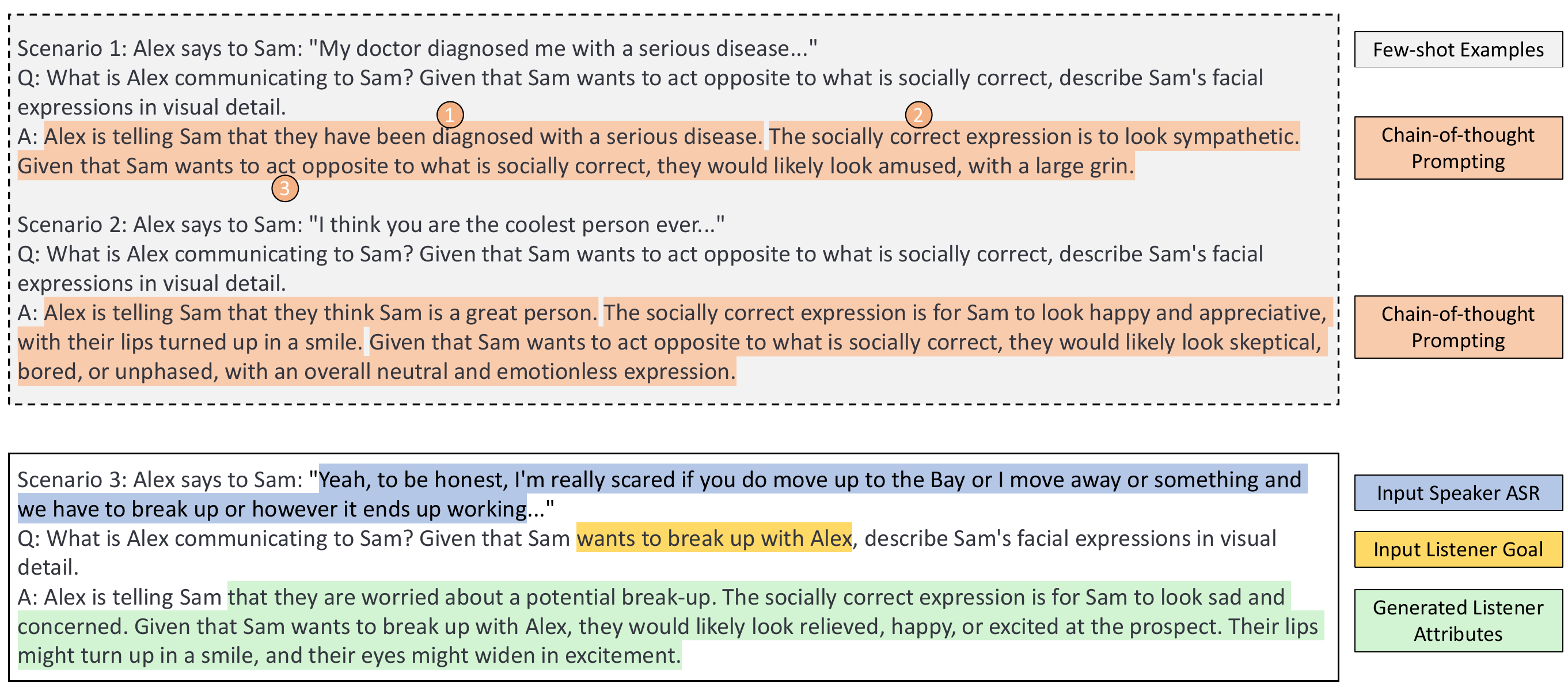}
    \vspace{-2em}
    \caption{\textbf{GPT-3 Few-shot Prompt:} We use chain-of-thought prompting~\cite{wei_chain_2022} and provide GPT-3 with two input-output examples following the desired reasoning process, traced by the orange circles in the figure. Specifically, we ask GPT-3 to (1) reason about what the speaker is saying, (2) reason about what a normal listener response might look like, and (3) reason about how the listener's goal might modulate their response. Given these few-shot examples, GPT-3 is able to make successful generations for unseen speakers and unseen listener goals.}
    \label{fig:gpt_prompt}
\end{figure*}

To address these issues, we provide GPT-3 with a few-shot prompt containing examples of the desired response format, as well as the desired reasoning structure~\cite{wei_chain_2022}. We present our prompt in Figure~\ref{fig:gpt_prompt}, which we use throughout our experiments.

\textbf{Generating Prompt Completions:} We generate completions using the public OpenAI API. We use the ``text-davinci-002" GPT-3 model, the largest publicly available model. For diversity in the generated listener attributes, we set the temperature parameter high ($=0.8$). We also set the ``max\_tokens" parameter high ($=1000$) to not to restrict the generated output length. All other parameters are kept at the default values suggested by the OpenAI API.

\subsection{Visual Prompt Details}
We use a combination of image-level prompting~\cite{bahng_exploring_2022} and token-level prompting~\cite{jia_visual_2022}. For image-level prompting, we pad input images with a fixed, randomly initialized,  learnable padding prompt of size 30, as described in ~\cite{bahng_exploring_2022}. For token-level prompting, we use Shallow Visual-Prompt Tuning~\cite{jia_visual_2022} and append 10 learnable tokens to the tokenized sequence of the input image. 

Overall, our setup allows us to train only a small number of parameters while keeping CLIP's large pretrained image and text encoder frozen. Specifically, there are $2Cp(H + W - 2p)$ learnable parameters for padding prompt, where $C, H, W, p$ are the image channels, height, width and padding size respectively. Since we use 224x224 size RGB images, we train $2 \times 3 \times 30 \times (224+224 - 2 \times 30) = 69840$ padding prompt parameters. For the token prompt, we learn $10 \times 768 = 7680$ parameters where $10$ is the number of tokens learned and 768 is the input dimension of CLIP ViT-B/32. Due to the low total parameter count, our training converged in a single epoch, taking less than 20 real-time seconds on a single 12GB GPU.

\section{Ethics Statement}
The RealTalk Dataset is constructed from publicly-available YouTube videos of real people who have consented to be filmed discussing their lives. We release it to spur research on social interaction. In doing so, we do not condone any misuse of our data in any manner that may be harmful to the people portrayed. We do not condone using our framework to misappropriate listener identities or to construe fake conversations.